\documentclass[letterpaper, 10 pt, conference]{ieeeconf}  

\IEEEoverridecommandlockouts                              

\usepackage{amsmath} 
\usepackage{amssymb}  
\usepackage[T1]{fontenc}
\usepackage[pdftex]{graphicx}
\usepackage{todonotes}
\graphicspath{{images/}}
\usepackage{multirow}
\usepackage{algorithm}
\usepackage[noend]{algpseudocode}
\usepackage{url}
\usepackage{graphicx}
\usepackage{tabularx}
\usepackage[colorlinks=false]{hyperref}
\usepackage[per-mode=symbol]{siunitx}

\usepackage{balance}
\usepackage{afterpage}

\newcommand{\rev}[1]{\textcolor{black}{#1}}

\newcommand{\norm}[1]{\left\lVert#1\right\rVert}
\DeclareMathOperator{\Tr}{Tr}

\DeclareMathOperator*{\argmin}{arg\,min}

\usepackage{placeins}
\usepackage[verbose]{wrapfig}

\makeatletter
\newcommand{\removelatexerror}{\let\@latex@error\@gobble}
\makeatother

\title{\LARGE \bf
 Occlusion-robust Deformable Object Tracking without Physics Simulation
}

\author{Cheng Chi$^{1}$ and Dmitry Berenson$^{1}$
\thanks{$^{1}$ Cheng Chi and Dmitry Berenson are with the University of Michigan, Ann Arbor, MI, USA.
        {\tt\small \{chicheng, dmitryb\}@umich.edu}. This work was supported in part by NSF grants IIS-1656101 and IIS-1750489 and ONR grant N000141712050.}%
}

\begin{document}

\maketitle
\thispagestyle{empty}
\pagestyle{empty}

\begin{abstract}

Estimating the state of a deformable object is crucial for robotic manipulation, yet accurate tracking is challenging when the object is partially-occluded. To address this problem, we propose an occlusion-robust RGBD sequence tracking framework based on Coherent Point Drift (CPD). To mitigate the effects of occlusion, our method 1) Uses a combination of locally linear embedding and constrained optimization to regularize the output of CPD, thus enforcing topological consistency when occlusions create disconnected pieces of the object; 2) Reasons about the free-space visible by an RGBD sensor to better estimate the prior on point location and to detect tracking failures during occlusion; and 3) Uses shape descriptors to find the most relevant previous state of the object to use for tracking after a severe occlusion. Our method does not rely on physics simulation or a physical model of the object, which can be difficult to obtain in unstructured environments. Despite having no physical model, our experiments demonstrate that our method achieves improved accuracy in the presence of occlusion as compared to a physics-based CPD method while maintaining adequate run-time.

\end{abstract}

\vspace{-0.05in}
\section{Introduction}
\vspace{-0.02in}

\begin{figure}
    \centering
    \vspace{4pt}
    \includegraphics[trim={0 0.5in 0 0},clip,width=.32\linewidth]{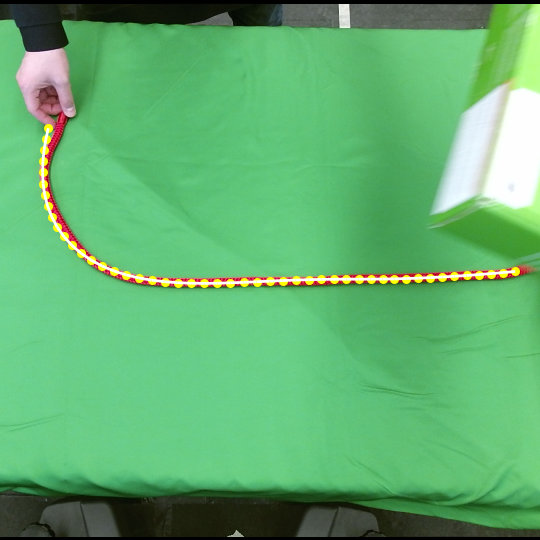}
    \includegraphics[trim={0 0.5in 0 0},clip,width=.32\linewidth]{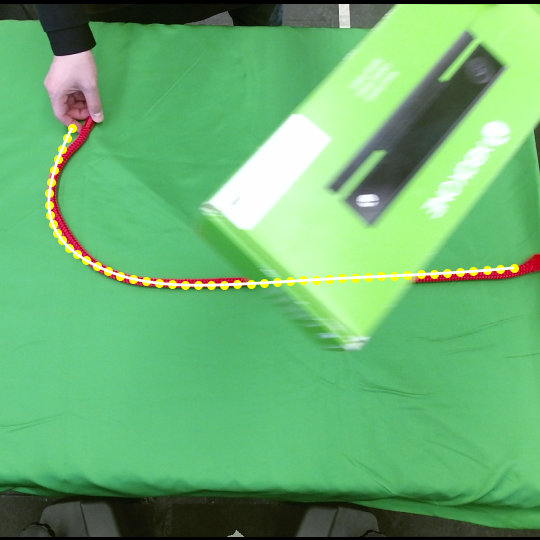}
    \includegraphics[trim={0 0.5in 0 0},clip,width=.32\linewidth]{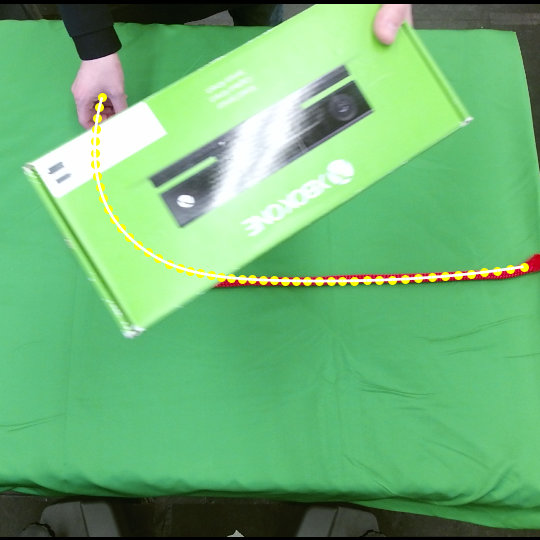}\\
    \vspace{0.05in}
    \includegraphics[width=.32\linewidth]{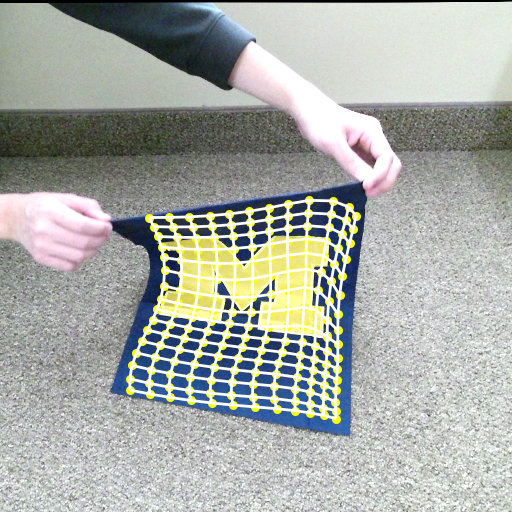}
    \includegraphics[width=.32\linewidth]{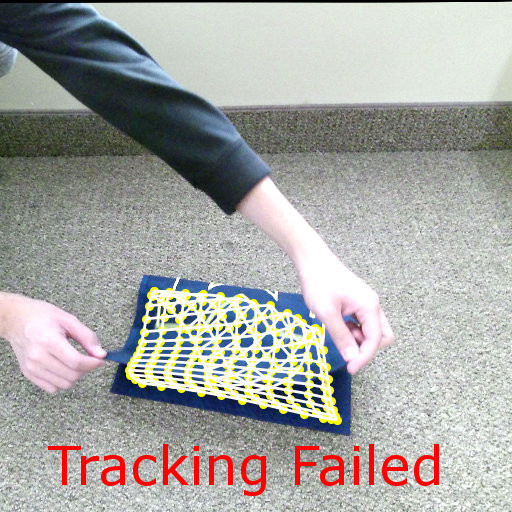}
    \includegraphics[width=.32\linewidth]{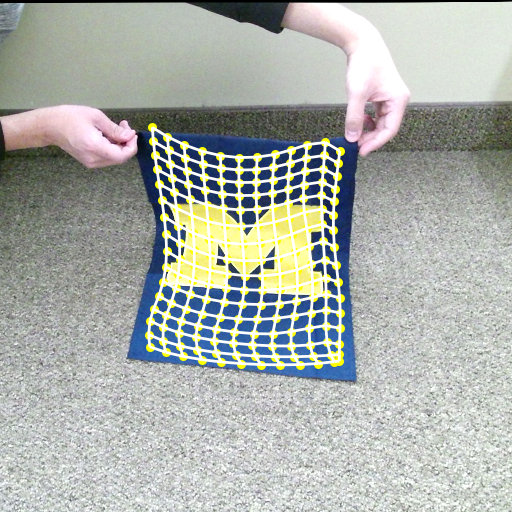}
    \caption{Illustration of tracking results for deformable objects. Yellow dots and white lines represent vertices and edges of the tracked model. Top: Tracking a rope under occlusion with our method. Time proceeds left to right. Bottom: Our method recovering from a tracking failure caused by self-occlusion.}
    \label{fig:intro}
\end{figure}

Tracking the geometry of deformable objects such as rope and cloth is difficult due to the continuous nature of the object (i.e. an infinite number of degrees of freedom), the lack of knowledge of the physical parameters of the object, and often a lack of visual features to track. In addition to these challenges, to be useful for robotic manipulation of deformable objects, a tracking algorithm must operate within a small computational budget and must be able to handle object self-occlusion (e.g. folding) and occlusion by objects in the environment (including the robot itself) (see Fig. \ref{fig:intro}). This paper thus focuses on producing an accurate estimate of the deformable object geometry during and after occlusion.

Tracking of deformable objects has been studied in the graphics \cite{graphics:White} \cite{graphics:Li} \cite{graphics:MsFusion}, computer vision \cite{cv:DynamicFusion} \cite{cv:Motion2Fusion} \cite{ml:CPD} \cite{cv:GLTP}, surgical \cite{surgical:Collins} \cite{surgical:heterogeneous}, and robotics \cite{robot:Petit} \cite{robot:TangState} fields. Most relevant to our domain is the work of \cite{robot:TangState}, which uses Coherent Point Drift (CPD) \cite{ml:CPD} to track the movement of points on the object and post-processes the output with a physics simulator to ensure that the predictions are physically-plausible. When an accurate physical model of the deformable object (including friction parameters) and all geometry in the environment is known, this method can be very effective. However, we seek a tracking method that does not require knowing this information in order to make deformable object tracking practical for robotic manipulation in unstructured environments such as homes, where physical and geometric models of obstacles may not be available.

Our method extends the literature on using CPD for deformable object tracking, making the following contributions to better handle occlusion: 1) We use a combination of locally linear embedding and constrained optimization to regularize the output of CPD, thus enforcing topological consistency when occlusions create disconnected pieces of the object; 2) We reason about the free-space visible by an RGBD sensor to better estimate the prior on point location and to detect tracking failures during occlusion; and 3) We use shape descriptors to find the most relevant previous state of the object to use for tracking after a severe occlusion.

In our experiments we compare our method with CPD \cite{ml:CPD} and CPD+Physics \cite{robot:TangState} for tracking rope and cloth manipulated by a human or robot.  Our results on both simulated and real data suggest that our method provides better accuracy than \cite{ml:CPD} and \cite{robot:TangState} when tracking deformable objects in the presence of occlusion while maintaining a fast run-time.

\vspace{-0.06in}
\section{Related Works}
\vspace{-0.04in}
Driven by the need to track human bodies and facial expressions, the graphics community has developed several related methods. White et al. \cite{graphics:White} demonstrated tracking fine motion of a cloth. However, this requires a color-coded cloth and inverse kinematics of the mesh. Li et al. \cite{graphics:Li} 
and Zollhofer et al. \cite{graphics:MsFusion} developed purely geometric methods of tracking arbitrary deformable objects. 
However, these methods are not robust against occlusion.

Computer vision researchers have shown promising results for deformable object reconstruction \cite{cv:DynamicFusion}, \cite{cv:Motion2Fusion}. By performing reconstruction in real-time, these methods avoid the occlusion problem by dynamically changing the tracked model. 
However, these methods do not satisfy the model consistency assumption required by the visual-servoing algorithms used for deformable object manipulation \cite{robot:Diminishing} \cite{robot:Navarro} \cite{robot:Dale}.

Researchers in surgical robots are also interested in deformable object tracking. \cite{surgical:Collins} and \cite{surgical:heterogeneous} have shown impressive results for tracking soft tissues in surgical scenarios, but these methods are specialized to the surgical domain and may not generalize to more flexible objects such as rope or cloth.

Prior work in robotics uses physics simulation to track deformable objects in the presence of occlusions \cite{robot:Schulman} \cite{robot:Petit}. A more recent approach \cite{robot:TangState} \cite{robot:TangControl} also uses Gaussian Mixture Model (GMM) and CPD algorithms to produce the input for its physics simulation engine. However, these algorithms must be given the physical model of the deformable object and the environment geometry, which we do not assume is available. However, as a baseline, we compare our method to \cite{robot:TangState} in our experiments.

\vspace{-.05in}
\section{Problem Statement}
\begin{figure}[t]
    \centering
    \vspace{4pt}
    \includegraphics[width=.47\linewidth]{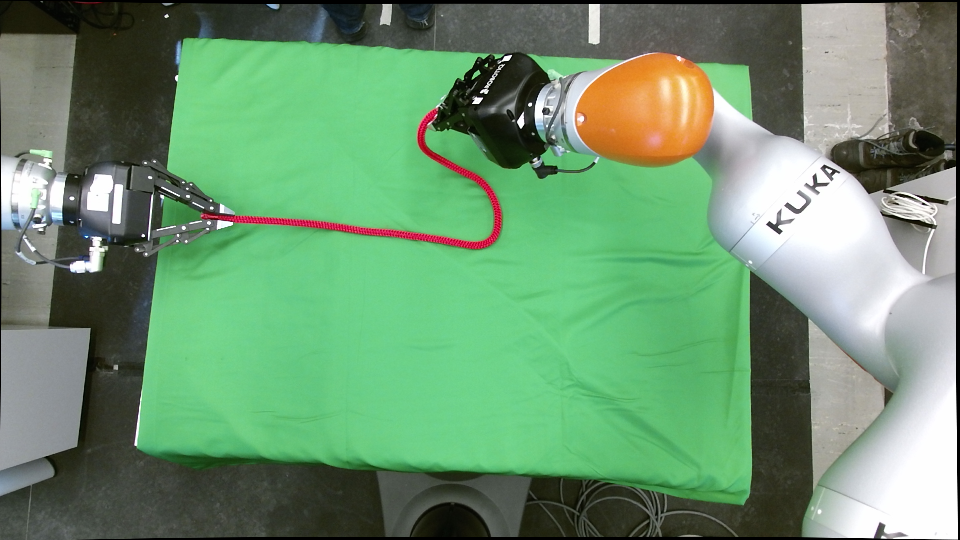} \includegraphics[width=.47\linewidth]{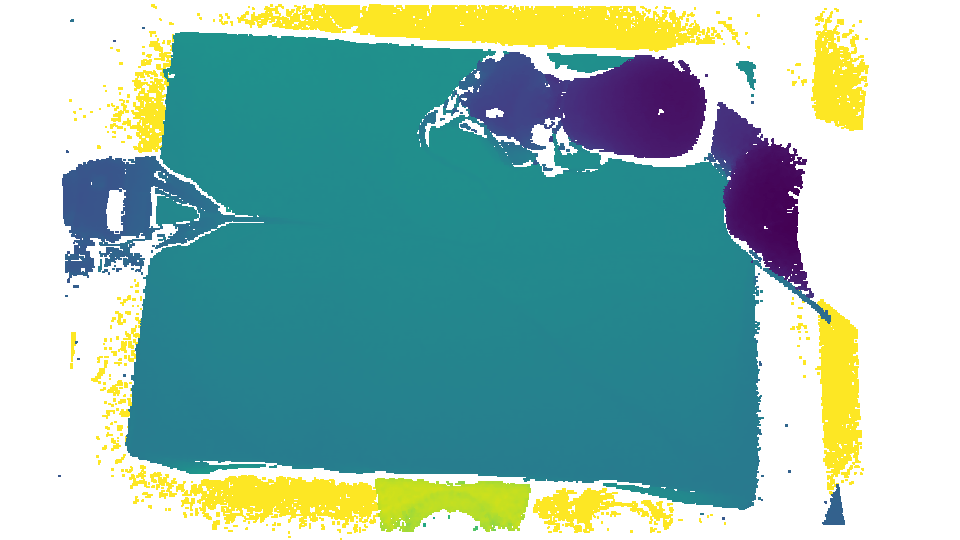} \\
    \vspace{0.05in}
    \includegraphics[width=.47\linewidth]{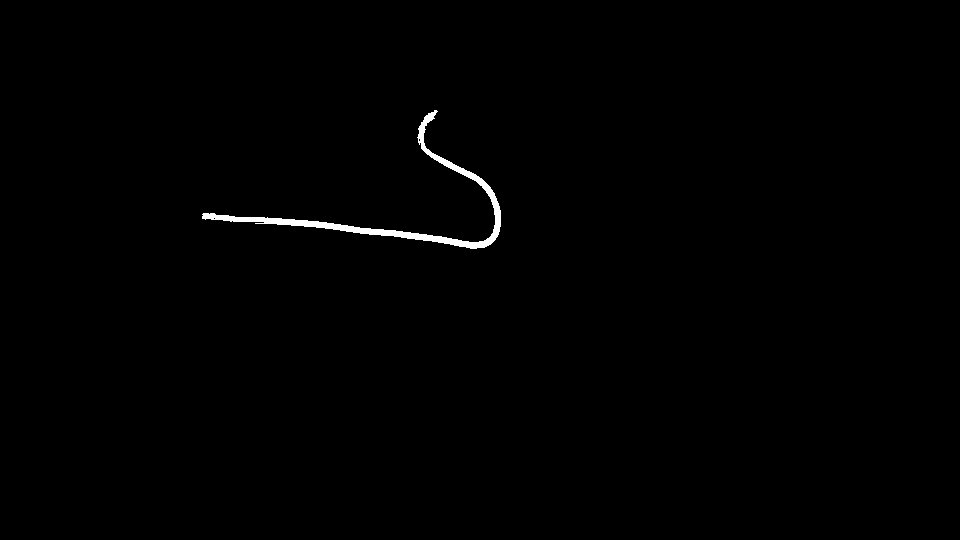} \includegraphics[width=.47\linewidth]{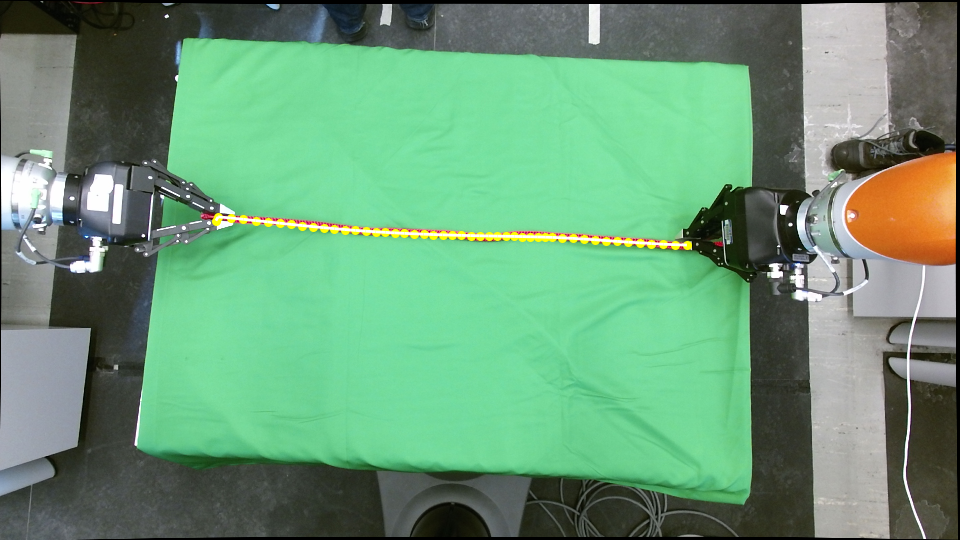} \\
    \caption{The input for each time step.  TL: RGB image. TR: depth image. BL: Mask image. BR: The input model for first time step, where yellow dots are vertices and white lines are edges.}
    \label{fig:input}
\end{figure}

\vspace{-.02in}
We define the problem of deformable object tracking as follows: Given a sequence of RGBD images $\mathcal{I}^t$ (Fig. \ref{fig:input}\rev{TL,TR}) for each time step $t$, also called a frame, and a connectivity model of the object with respect to the first RGBD image (Fig. \ref{fig:input}\rev{BR}) $<Y^0, E>$, where $Y^0=[y^0_1,y^0_2,\dots,y^0_M]^T \in \mathbb{R}^{M\times D}$ is a set of vertices and $E=[e_{0}, e_{1},\dots,e_{K}]^T\in \mathbb{I}^{K \times 2}$ is a set of edges; each $e_k \in E$ holds two integer indices $i,j$ of $Y^0$, representing an edge between vertices $y^0_i$ and $y^0_j$. Our goal is to produce a state $Y^t$ for each time step 
that is consistent with the geometry of the deformable object.
While segmenting pixels in an RGBD image that belong to our object of interest is an important step to solve this problem, it is beyond the scope of this paper, so we also assume that each RGBD image has an associated object mask $\mathcal{M}^t$ with the same size as $\mathcal{I}^t$ (Fig. \ref{fig:input}\rev{BL}). Mask images have value $1$ for all pixels belonging to the object of interest, and $0$ otherwise. Using $\mathcal{I}^t$ and $\mathcal{M}^t$, along with the camera parameters, we can convert the input to a point cloud $X^t=[x^t_1,x^t_2,\dots,x^t_N]^T\in \mathbb{R}^{N\times D}$.
In our case, $D=3$. We assume the topology of the object remains constant so we can represent the state of the object at time $t$ simply as $Y^t$, where $y^t_i$ and $y^{t'}_i$ correspond to the same point on the object for frames $t$ and $t'$.

If we denote the true state of the object at time $t$ as $Y^{*t}$, then our output should minimize $\norm{Y^{*t} - Y^t}_2$. 
The high-dimensional state space combined with the difficulty of establishing correspondences between frames makes dealing with occlusions especially difficult. Later in this paper, we will discuss when the assumption that the provided $Y^0$ corresponds to the true initial state is violated. We will also discuss how to leverage additional knowledge when prior correspondence, such as robot gripping points, are available.

\vspace{-0.05in}
\section{Tracking Deformable Objects}
\vspace{-0.02in}
Our proposed algorithm for deformable object tracking first pre-processes the given model $<Y^0, E>$. Then, for each $\mathcal{I}^t$ and $\mathcal{M}^t$ we generate $X^t$ and perform registration between $X^t$ and $Y^{t-1}$ using a Gaussian Mixture Model (GMM) Expectation-Maximization (EM) (\S\ref{sec:gmm}). We enhance GMM EM to better handle occlusions by using a novel visibility prior (\S\ref{sec:visprior}). The GMM EM is then regularized using existing methods \cite{ml:CPD} \cite{cv:GLTP} (\S\ref{sec:cpd}, \S\ref{sec:lle}).  The output of the GMM EM loop is adjusted by our proposed constrained optimization approach, in which we enforce stretching constraints and can incorporate known correspondences, such as robot gripping points (\S\ref{sec:constrained}). Finally, a novel tracking failure detection and recovery algorithm is executed to recover tracking after a large occlusion (if necessary) (\S\ref{sec:recovery}). The overall algorithm is shown in Algorithm~\ref{alg:AllAlgorithm}.

\vspace{-.05in}
\subsection{GMM-Based point Set Registration}
\vspace{-.02in}
\label{sec:gmm}
Between two consecutive time steps $t$ and $t+1$, the change in appearance of the deformable object is relatively small. Thus, it is reasonable to formulate this frame-to-frame tracking problem as a point-set registration problem, i.e. to estimate $Y^{t+1}$ by aligning $Y^{t}$ to $X^{t+1}$. Following the formulation in \cite{ml:CPD} \cite{cv:GLTP}, we will formulate this point-set registration problem using a GMM. For readability, the rest of this section will use $Y$ instead of $Y^t$, $Y'$ instead of $Y^{t+1}$ and $X$ instead of $X^{t+1}$. We consider each $x_{n} \in X$ as a sample generated from a mixture of independently distributed Gaussian distributions, where each $y_m \in Y$ is the mean of a Gaussian distribution. We assume these Gaussian distributions share the same isotropic variance $\sigma^2$ and membership probability of $\frac{1}{M}$. Thus, the probability distribution of point $x_n$ can be written as:
\begin{align}
    p(x_n)=\sum^M_{m=1}\frac{1}{M}\frac{1}{(2\pi \sigma^2)^{\frac{D}{2}}}\exp\left(-\frac{\norm{x_n-y_m}^2}{2\sigma^2}\right)
\end{align}

To account for noise and outliers in $X$, a uniformly distributed component with index $M+1$ is added to the mixture model with weight $\omega, 0\leq \omega \leq 1$. Thus, the joint probability density function of our GMM can be written as:
\begin{align}
    p(X)=\prod^N_{n=1}p(x_n)=\prod^N_{n=1}\sum^{M+1}_{m=1}p(m)p(x_n|m) \label{eqn:GmmPdf}
\end{align}
where
\begin{align}
    p(x | m) &=
    \begin{cases}
        \frac{1}{(2\pi \sigma^2)\frac{D}{2}}\exp\left(-\frac{\norm{x-y_m}}{2\sigma^2}\right),& m=1,\dots,M\\
        \frac{1}{N},& m=M+1
    \end{cases}\\
    p(m) &=
    \begin{cases}
        \frac{1-\omega}{M}, & m=1,\dots,M\\
        \omega, & m=M+1
    \end{cases}
\end{align}
Then, our goal is to find the $Y'$ and $\sigma^2$ that maximizes the log-likelihood of Eq.~\ref{eqn:GmmPdf}, with values of $Y'$ initialized from $Y$. 
We can solve this problem following the Expectation Maximization algorithm described in \cite{ml:CPD} \cite{cv:GLTP} \cite{robot:TangState}. In the E-step, we find the posterior probabilities using the current GMM parameters based on Bayes rule:
\begin{align}
    p^{cur}(m|x_n)=
    \frac{
        \exp\left(-\frac{1}{2}\norm{\frac{x_n - y_m}{\sigma}}^2\right)
        }{
        \sum^M_{i=1}\exp\left(-\frac{1}{2}\norm{\frac{x_n - y_m}{\sigma}}^2\right) + \frac{(2\pi\sigma^2) \frac{D}{2}\omega M}{(1-\omega) N}
        }
\end{align}
Then, in the M-step, we obtain the optimal $Y'$ and $\sigma^2$ by minimizing the following cost function:
{\small
\begin{align}
\begin{split}
    & Q(y_m, \sigma^2)=\\
    & -\sum^N_{n=1}\sum^{M}_{m=1}p^{cur}(m|x_n)\left(\log\left(\frac{1-\omega}{M(2\pi\sigma^2)^{\frac{D}{2}}}\right)-\frac{\norm{x_n - y_m}^2}{2\sigma^2}\right)\\
    & -\sum^N_{n=1}p(M+1|x_n)\log\left(\frac{\omega}{N}\right) \label{eqn:OrigQ}
\end{split}
\end{align}
}
We iterate the E- and M-steps until convergence.

However, we found that the above algorithm is not sufficient for solving the deformable object tracking problem. When parts of the object are occluded, the GMM EM algorithm will overfit to visible points, breaking smoothness and local topology properties of the original model (Fig. \ref{fig:gmm}\rev{L}). This problem suggests that proper regularization is needed.

\vspace{-.05in}
\subsection{Addressing Occlusion by Exploiting Visibility Information}
\begin{figure}
    \centering
    \vspace{4pt}
    \includegraphics[width=0.48\linewidth]{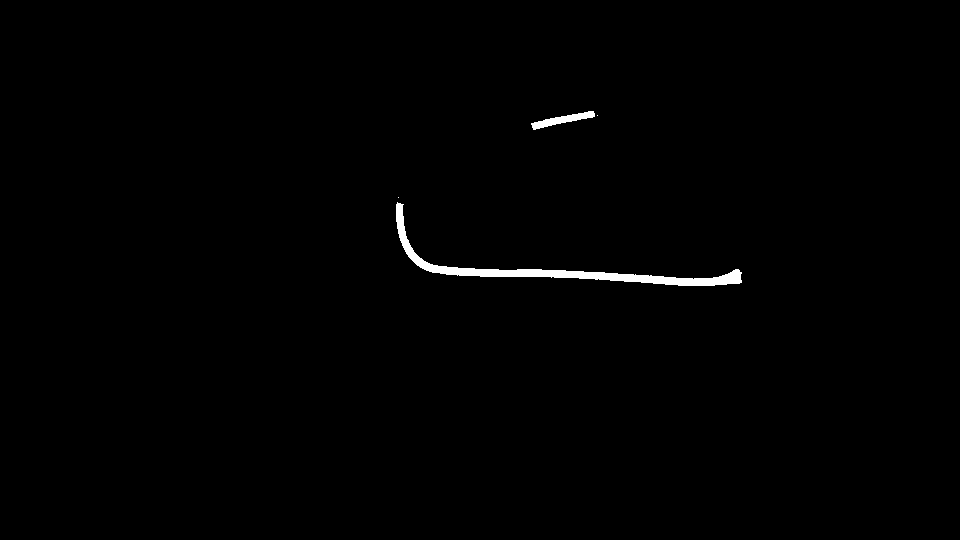}\hfill%
    \includegraphics[width=0.48\linewidth]{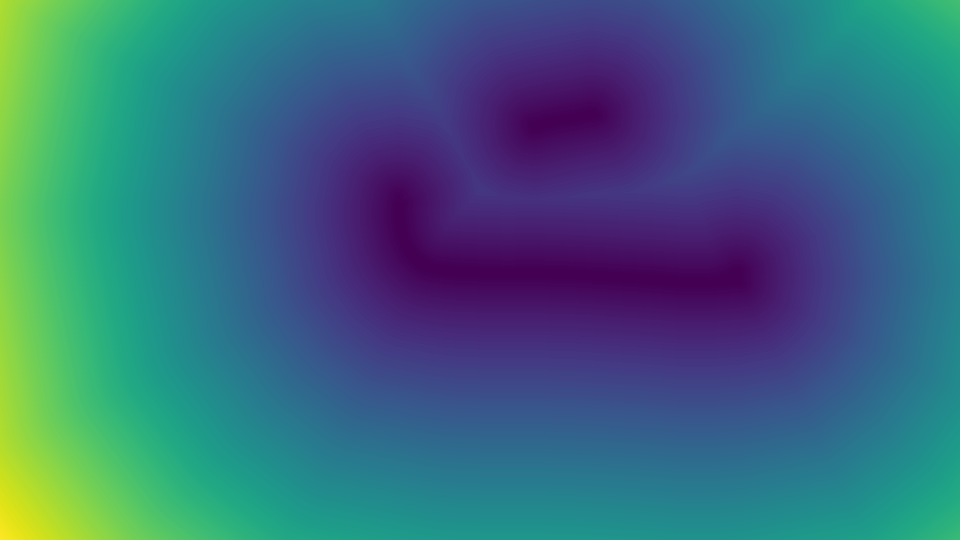}
    \caption{Distance image (right) generated by L2 distance transform of mask image (left).}
    \label{fig:dist}
\end{figure}

\vspace{-.02in}
\label{sec:visprior}
Previously, we assumed that the membership probabilities of all Gaussian distributions are $p(m)=\frac{1}{M}$, i.e. it is equally possible for each Gaussian distribution to generate an $x_n$. In practice, especially under occlusion, we found the above assumption does not hold. For example, if we knew with certainty that $y^t_m$ had been occluded at time $t$, we should not expect to observe any sample generated from $y^t_m$, i.e. $p(m^t)=0$. While it is impossible to directly derive $p(m^t)$ without knowing $y^t_m$, we can still approximate $p(m^t)$ using $y^{t-1}_m$ and visibility information. Since the movement of the deformable object is usually small between consecutive frames, if $y^{t-1}_m$ is occluded in $\mathcal{I}^t$, it should be less likely to generate observation samples than other Gaussian distributions. 
Let $\mathcal{I}_d^t(u, v)$ denote the depth value at the $u, v$ location of $\mathcal{I}^t$. Let $\mathcal{D}^t$ denote the image generated by a distance transform of $\mathcal{M}^t$ (Fig. \ref{fig:dist}), where the value of each pixel in $\mathcal{D}^t$ denotes the Euclidean distance from $(u,v)$ to the nearest pixel in $\mathcal{M}^t$ that has value $1$.
For any $y^{t-1}_m$, we can obtain its coordinates in the image coordinate frame $(u_m^{t-1},v_m^{t-1})$ and depth value $z_m^{t-1}$ using camera parameters. We can then approximate $p(m^t)$ as:
\begin{align}
\begin{split}
    & p_{vis}(m^t)=\mu_{vis} e^{-k_{vis}\mathcal{D}^t(u_m^{t-1},v_m^{t-1})\cdot\max(z_m^{t-1}-\mathcal{I}_d^t(u_m^{t-1},v_m^{t-1}), 0)}
\label{eqn:Pvis}
\end{split}
\end{align}

\afterpage{
\clearpage\clearpage
    \begin{figure}[t]
        \centering
        \vspace{4pt}
        \includegraphics[width=0.49\linewidth]{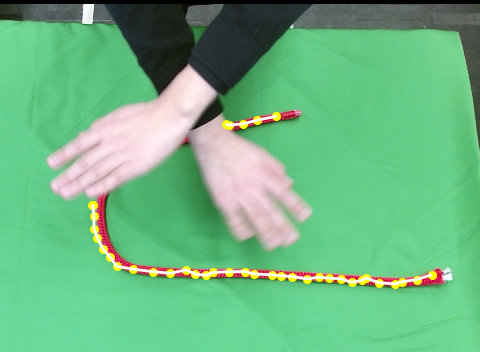}\hfill%
        \includegraphics[width=0.49\linewidth]{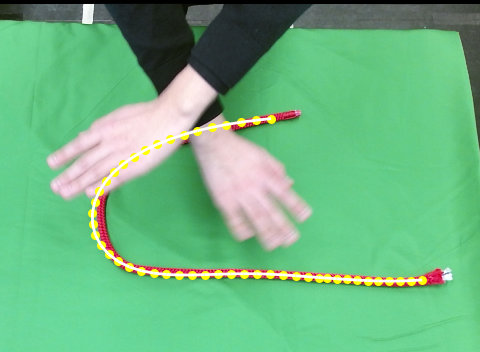}\\
        \caption{Left: occluded scene tracked with GMM. Right: tracking result of GMM with CPD and LLE regularization.}
        \label{fig:gmm}
    \end{figure}
    \noindent
    \vspace{-0.25in}
    \begin{algorithm}[H]
        \caption{Track$(X^t,Y^{t-1}, H, Y^0, E)$}
        \begin{algorithmic}[1]
            \State Compute $p_{vis}(m^{t})$ for all $y^{t-1}_m \in Y^{t-1}$ using Eq.~\ref{eqn:Pvis}
            \State $\sigma^2 \gets Var(X^t)$
            \State $W \gets 0$
            \While{$\sigma^2 > \epsilon$}
                \State Compute $P$ using Eq.~\ref{eqn:GmmPdf}
                \State Compute $G$ using Eq.~\ref{eqn:DefineSpatialTransform}
                \State Solve for $W$ using Eq.~\ref{eqn:SolveW}
                \State Compute new $\sigma^2$ using Eq.~\ref{eqn:SolveSigma}
            \EndWhile
            \State $Y^t \gets Y^{t-1}+GW$
            \State Solve for $Y^{*t}$ using Eq.~\ref{eqn:QCQPGripper}
            \State \Return $Y^{*t}$
        \end{algorithmic}
        \label{alg:track}
    \end{algorithm}
    \vspace{-0.2in}
    \begin{algorithm}[H]
        \caption{Main Loop}
        \begin{algorithmic}[1]
            \State \textbf{Data:} $Y^0$, Vertices of the initial model
            \State \textbf{Data:} $E$, Edges of the initial model
            \State $F \gets \emptyset$, Set of VFH shape descriptors
            \For{$i \in \{1,2,\dots, \}$}
                \State \textbf{Input:} $\mathcal{I}^t, \mathcal{M}^t$
                \State Compute $X^t$ from $\mathcal{I}^t, \mathcal{M}^t$
                \State $Y^t\gets \text{Track}(X^t,Y^{t-1},H,Y^0,E)$
                \State Compute $J_{free}(Y^t)$ with Eq.~\ref{eqn:Jfree}
                \State $f(X^t)\gets$ VFH shape descriptor of $X^t$
                \If{$J_{free}(Y^t)<\tau$}
                    \State $F \gets F \cup f(X^t)$
                    \State \textbf{Output:} $Y^t$
                \Else
                    \State $K\gets$ Query kNN of $f(X^t)$ in $F$
                    \For{$k$ in $K$}
                        \State $Y'^k\gets \text{Track}(X^t,Y^{k}, H, Y^0, E)$
                        \State Compute $J_{free}(Y'^k)$ using Eq.~\ref{eqn:Jfree}
                    \EndFor
                    \State $Y^t\gets Y'^k$ with minimum $J_{free}(Y'^k)$
                    \State \textbf{Output:} $Y^t$
                \EndIf
            \EndFor
        \end{algorithmic}
        \label{alg:AllAlgorithm}
    \end{algorithm}
    \vspace{-0.12in}
}
\noindent
where $\mu_{vis}$ is a normalization factor such that $\sum^M_{m=1}p_{vis}(m^t)=1$ and $k_{vis}$ is a parameter that controls the influence of visibility information. We formulate Eq.~\ref{eqn:Pvis} so that vertices below the visible point cloud are more likely to be penalized (the $\max(\cdot)$ term), and points farther away from the object are more likely to be penalized (the $\mathcal{D}^t(\cdot)$ term). We intentionaly do not consider self-occlusion in this estimate because inaccurate estimation of $Y^{t-1}$ created false positives in practice. 
Combining this formula with our uniform distribution component, we get:
\begin{align}
    p(m^t)=
        \begin{cases}
        (1-\omega)p_{vis}(m^t), & m=1,\dots,M\\
        \omega, & m=M+1
    \end{cases}
\end{align}
We can then derive our new expression for posterior probabilities in the E-step:
\begin{multline}
    p^{cur}(m|x_n)= \\
    \frac{p_{vis}(m)\exp\left(-\frac{1}{2}\norm{\frac{x_n-y_m}{\sigma}}^2\right)}
    {\sum^M_{i=1}p_{vis}(m)\exp\left(-\frac{1}{2}\norm{\frac{x_n-y_m}{\sigma}}^2\right) + \frac{(2\pi\sigma^{2})^{\frac{D}{2}}\omega}{(1-\omega)N}}
    \label{eqn:OurPOld}
\end{multline}

\vspace{-.05in}
\subsection{Coherent Point Drift}
\vspace{-.02in}
\label{sec:cpd}
In the aforementioned GMM registration method, we assume that all Gaussian distributions are independent. However, this assumption is not accurate in our setting. For a deformable object, we can often observe that points residing near each other tend to move similarly \cite{ml:CPD}. We thus use the Coherent Point Drift (CPD) \cite{ml:CPD} regularization which preserves local motion coherence to better represent the true motion. Instead of modeling each point $\rev{y}^t_m$ as an independently moving Gaussian centroid, CPD embeds the frame-to-frame change in a spatial transformation $y^t_m = \mathcal{T} (y^{t-1}_m, W^t)$ that maps every point in the space around our object of interest at time $t-1$ to another point at time $t$ using parameter matrix $W^t \in \mathbb{R}^{M\times D}$. 

More specifically, CPD represents this spatial transformation as a Gaussian Radial Basis Function Network (GRBFN):
\begin{align}
    \mathcal{T}(y^{t-1}_m,W^t) &=y^{t-1}_m + v(y^{t-1}_m)\\
    v(z) &=\sum^M_{m'=1}w^t_{m'}g(z-y^{t-1}_{m'})\\
    g(z-y^{t-1}_{m'}) &=\exp\left(-\frac{\norm{z-y^{t-1}_{m'}}^2}{2\beta^2}\right)
\end{align}
Where $w^t_{m'}\in\mathbb{R}^D$ is the $m'$th row of $W^t$. In this deformation field $v$, for every $z\in \mathbb{R}^3$, $v(z)$ output a vector that represent the displacement for position $z$. GRBFN has several desirable properties: it is smooth and differentiable everywhere, and is linear except for the radial basis function itself. Here, $\beta$ is a hyper-parameter that controls the widths of Gaussian kernels, which affects the rigidity of the deformation field. CPD puts centroids of Gaussian kernels at every $y^t_m$. We can represent the spatial transformation in matrix-vector form:
\begin{align}
    Y^{t}=\mathcal{T} (Y^{t-1}, W^t)=Y^{t-1} + G^{t-1}W^t
    \label{eqn:DefineSpatialTransform}
\end{align}
where $G^{t-1}_{M\times M}$ is a Guassian kernel matrix with element $G^{t-1}_{ij}=\exp\left(-\frac{1}{2 \beta^2} \norm{y^{t-1}_i-y^{t-1}_j}^2 \right)$. We can then regularize the weight matrix $W$ to enforce the motion coherence:
\begin{align}
    E_{CPD}(W)=\Tr{(W^T GW)}
\end{align}
$E_{CPD}$ will be used as a energy term in the M-step of our GMM EM algorithm, where $W$ will become the parameters to be optimized (shown below).

\vspace{-.05in}
\subsection{Preserving Topology using Locally Linear Embedding}
\vspace{-.02in}
\label{sec:lle}
In practice, we found that simply adding CPD regularization is not sufficient. Since CPD only enforces motion coherence between consecutive frames, the error between the shape of the current tracking state and the true state will accumulate. This will cause the topology of the tracking result to drift away from the original model, even though the tracking result is statistically correct when viewed as a point set registration problem. To solve this problem, we will use a regularization term proposed in \cite{cv:GLTP}, which is based on Locally Linear Embedding (LLE) \cite{ml:LLE}, to enforce topological consistency with respect to the original model.

LLE performs non-linear dimensional reduction while preserving local neighborhood structure. We found that the assumption of LL, that the data lies on a low dimensional manifold, holds true for our deformable object tracking setting. When tracking an object such as a rope, $X^t$ and $Y^t$ mostly reside on a 1D curve. When tracking a cloth, $X^t$ and $Y^t$ reside on a 2D surface. Thus, we will generate a LLE of our model $Y^0$ before the tracking starts, 
and use this embedding as part of our regularization. 

More specifically, we first represent every point in our model as a linear combination of its k-nearest neighbors. We can obtain linear weights $L$ by minimizing the following cost function:
\begin{align}
    J(L)=\sum^M_{m=1}\norm{y^0_m - \sum_{i\in K_m}L_{mi}y^0_i}^2
\end{align}
where $K_m$ is a set of indices for the $k$ nearest neighbors of ${\bf y}^0_m$, and $L$ is a $M\times M$ adjacency matrix where $L_{ij}$ represent a edge between $y^0_i$ and $y^0_j$ with their corresponding linear weight if $j \in K_{i}$ and $0$ otherwise.  We then define a regularization term that penalize the deviation from the original local linear relationship (for readability, we will drop the time index $t$ for subsequent appearance of $W$):
\begin{align}
\begin{split}
    E_{LLE}(W) &= \sum^M_{m=1}\norm{y^t_m-\sum^K_{i\in K_m}L_{mi}y^t_m}^2\\
    & =\sum^M_{m=1}\norm{\mathcal{T}(y^{t-1}_m, W)-\sum^K_{i\in K_m}L_{mi}\mathcal{T}(y^{t-1}_i, W)}^2
    \label{eqn:Elle}
\end{split}
\end{align}

Note that, unlike \cite{cv:GLTP}, we only compute $L$ once on $Y^0$ and use this matrix for all subsequent regularized GMM EM operations in the following frames. Replacing $y^t_m$ in Eq.~\ref{eqn:OrigQ} with the expression of $\mathcal{T} (y^{t-1}_m, W^t)$ and adding CPD and LLE regularization terms, we now have our new cost function for the M-step:
\begin{align}
    Q(W, \sigma^2)=
    \sum^{M, N}_{m,n=1} p^{cur}(m|x^t_n)\frac{
    \norm{x^t_n-(y^t_m + G(m,\cdot)W)}^2
    }{
    2\sigma^2
    }\nonumber\\
    +\frac{N_p D}{2}\ln{(\sigma^2)}
    +\frac{\alpha}{2}E_{CPD}(W)
    +\frac{\gamma}{2}E_{LLE}(W)
\end{align}
where $N_p=\sum^{N,M}_{n,m=1}p^{cur}(m|x_n)$, $\alpha$ and $\gamma$ are parameters that trade off between GMM matching, frame-to-frame motion coherence, and local topological consistency. Note that we removed terms in Eq.~\ref{eqn:OrigQ} that are independent of $W$ and $\sigma^2$, as we will only optimize with respect to these two variables.

Using the process described in \cite{cv:GLTP}, we can perform the E-step by first obtaining $W$ by solving a system of linear equations:
\begin{multline}
    (d(P{\bf 1})G +\sigma^2\alpha I + \sigma^2\lambda HG)W \\= PX - (d(P{\bf 1})+\sigma^2\lambda H)Y
    \label{eqn:SolveW}
\end{multline}
where each entry of $P\in\mathbb{R}^{M\times N}$ contains $p^{cur}(m|x^t_n)$, where ${\bf 1}$ is a column vector of ones, $I$ denotes the identity matrix, and $d(v)$ represent the diagonal matrix formed by vector $v$, and $H\in\mathbb{R}^{M\times M}=(I-L)^T(I-L)$.
We then obtain $\sigma^2$:
\begin{align}
\begin{split}
    \sigma^2 &=\frac{{\bf 1}}{N_p D}(\Tr{(X^T d(P^T{\bf 1})X)})-2\Tr{(Y^TPX)}\\
    &-2\Tr{(W^T G^T PX)} + \Tr{(Y^T d(P{\bf 1})Y\rev{)}}\\
    &+2\Tr{(W^T G^T d(P{\bf 1})Y)} + \Tr{(W^T G^T d(P{\bf 1})GW)}
    \label{eqn:SolveSigma}
\end{split}
\end{align}
After the EM algorithm converges, we set $Y^{t+1}=Y^{t}+G^{t}W^{t}$. We can see the improvement as compared to the original GMM algorithm in Fig. \ref{fig:gmm}\rev{R}.

\vspace{-.05in}
\subsection{Enforcing Stretching Limits via Constrained Optimization}
\begin{figure}[t]
    \centering
    \vspace{4pt}
    \includegraphics[width=0.49\linewidth]{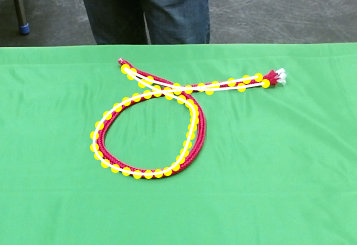}\hfill%
    \includegraphics[width=0.49\linewidth]{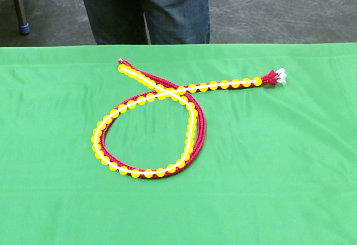}
    \caption{Left: tracking result of GMM with CPD and LLE regularization. Right: tracking result of GMM with CPD, LLE regularization, and constrained optimization post processing.}
    \label{fig:with_constraint}
\end{figure}

\vspace{-.02in}
\label{sec:constrained}
Within the GMM Expectation-Maximization loop, the added $E_{LLE}$ regularization term mitigated the topological consistency problem. However in practice, we found that $E_{LLE}$ is not sufficient to address anisotropic effects in deformable object motion. Many deformable objects, such as rope and cloth are less deformable when being stretched than being compressed or being bent. However, $E_{LLE}$, using the locally linear assumption, will treat stretching, compression, and bending almost the same. The artifact created by stretching can be seen in Fig. \ref{fig:with_constraint}. Thus, we introduced a constrained optimization method to post-process the output of the GMM EM loop so that the output allows bending and compression while keeping the distance between points below a threshold. Specifically,
\begin{align}
\begin{split}
    & Y^{*t} = \argmin_{Y^*}\sum^M_{m=1}\norm{Y^*_m - Y^t_m}^2\\
    & \text{subject to} \norm{Y^{*t}_i-Y^{*t}_j}\leq \lambda \norm{Y^{0}_i-Y^{0}_j}\:\: \forall (i,j) \in E
    \label{eqn:QCQP}
\end{split}
\end{align}
where $Y^{*t}_m$ is the $m$th row of $Y^{*t}$ and $\lambda \geq 1$ is a parameter that controls the flexibility of our constraints. This step restricts the length of any edge in the tracking result to be at most $\lambda$ times the length of the same edge in the original model. We used the Gurobi Package \cite{software:gurobi} to solve this optimization problem. Fig.~\ref{fig:with_constraint} shows the tracking result with constrained optimization added.

\subsubsection{Incorporating Prior Correspondence}
\label{sec:gripper}
When deformable object tracking is being used as part of a robotic manipulation system, we can often obtain partial but reliable knowledge about correspondence. For example, when a rope is being dragged by a robot gripper, we know exactly where the gripping point on the rope should be in 3D space given the robot's configuration and forward kinematics. Such information can be easily incorporated into our constrained optimization framework.

We represent a set of known correspondences as $<Z^t, C^t>$, where $Z^t=[z^t_1,z^t_2,\dots,z^t_K]^T\in \mathbb{R}^{K\times D}$ is a set of known points, and $C^t=[\rev{c_0}, \rev{c_1},\dots,c_K]^t\in \mathbb{I}^{K\times 2}$, where each $c\in C^t$ contain a pair of indices $m, k$ that represent a correspondence between a known point and a model vertex. Our constrained optimization formulation then becomes:
\begin{align}
\begin{split}
    & Y^{*t} = \argmin_{Y^*}\sum^M_{m=1}\norm{Y^*_m - Y^t_m}^2\\
    & \text{subject to} \norm{Y^{*t}_i-Y^{*t}_j}\leq \lambda \norm{Y^{0}_i-Y^{0}_j} \: \forall (i,j) \in E\\
    & \text{and }Y^{*t}_m=Z^t_k \:\: \forall (m,k) \in C^t
    \label{eqn:QCQPGripper}
\end{split}
\end{align}
Incorporating this constraint guarantees $Y^{*t}$ satisfies our known correspondences.

\vspace{-.05in}
\section{Tracking Failure Recovery}
\vspace{-.02in}
\label{sec:recovery}
The above methods allow us to track the object with moderate occlusion. However, there are cases where tracking becomes impossible. Imagine the object is temporarily completely blocked by something in front of the RGBD camera, while the object itself moves. In this case, we have no information with which to infer the state of the object, and tracking will fail. However, a problem arises when the occlusion disappears. Since we initialize tracking and regularize deformation from the last time step, if the object deformed too much while being completely occluded, we might never be able to track correctly again. Thus, we propose a novel tracking failure recovery system. 

\vspace{-.05in}
\subsection{Tracking Failure Detection}
\vspace{-.02in}
Since there will be an additional computational cost associated with tracking failure recovery, we will first detect whether a tracking failure has occured, and only apply recovery when needed. We will infer tracking failure based on free-space information: For each ray emanating from the camera, we know that the space between 0 to the depth value of the visible point along that ray will not contain another visible object. Thus, if vertex of $Y^t$ lies in that space, we know the position for that vertex is wrong. Similar to our visibility prior (Eq.~\ref{eqn:Pvis}), we construct an energy function that indicates the percentage of vertices that are in free space and how far away they are from non-free space:
\begin{align}
\begin{split}
    & J_{free}(Y^t)=\\
    & \frac{1}{M}\sum^M_{m=1}e^{-k\mathcal{D}^t(u^t_m,v^t_m)\max(\mathcal{I}_d^t(u^t_m,v^t_m)-z^t_m, 0)}
    \label{eqn:Jfree}
\end{split}
\end{align}

When $J_{free}(Y^t) > \tau$, we will treat $Y^t$ as failing to track, and invoke tracking failure recovery (described below). $\tau$ is a threshold we set manually.

\vspace{-.05in}
\subsection{kNN Template retry}
\vspace{-.02in}
We assume that the new state of the deformable object is similar to some state we have seen before and correctly tracked. This assumption might not always hold, but we found it often works in practice when tracking fails. If tracking fails at time $t'$, we re-initialize our tracking algorithm with a $Y^t$ where $t < t'$. Since our time and computation resources are limited, and the number of $Y^t$s grows linearly with time, we will only retry tracking on previous states that we think are the most similar to our current state. We will measure this similarity using a 3D shape descriptor: Viewpoint Feature Histogram (VFH) \cite{robot:Vfh}. VFH computes a histogram of the angle between viewpoint ray and object surface normals estimated from a point cloud. For each time $t$ where $J_{free}(Y^t)\leq\tau$ we compute a VFH descriptor from $X^t$, which is then stored in a library. When a tracking failure is detected at time $t'$, we compute another VFH descriptor from $X^{t'}$, and query the k nearest neighbors from our descriptor library, measured in Euclidean distance, using Fast Library for Approximate Nearest Neighbors (FLANN) \cite{ml:Flann}. We run all k neighbors through the tracking method and select the result with the lowest $J_{free}$.

The shape descriptor library will grow as tracking proceeds, which may creates two problems for long sequences. First, as the set of descriptors becomes more and more dense, the result of the kNN query results will become increasingly similar, and the output of each tracking result will also be similar. Second, the kNN query itself will become increasingly time-consuming. The sequences used in our experiments were not long, so we did not encounter these issues. However, for longer sequences it would be straightforward to cluster the shape descriptors using the k-means algorithm and only preserve a single shape descriptor from each cluster.

\vspace{-.05in}
\subsection{Tracking without true state in the first frame}
\vspace{-.02in}
For many robotic manipulation applications, we are not given the true state of the object at the first frame. It is useful to view this problem as a special case of tracking failure. We can use a shape descriptor library generated offline and perform the kNN retry routine described above. To obtain such a shape descriptor library, we can set up a training scenario where the object starts from a simple, known state. We can then manipulate the deformable object to obtain various other states and their shape descriptors, and perform k-means down-sampling before using the library for a new scene.

\vspace{-.05in}
\section{Results}
\vspace{-.02in}
\begin{figure*}
    \centering
    \vspace{4pt}
    \includegraphics[trim={0 2.85in 0 0},clip,width=\linewidth]{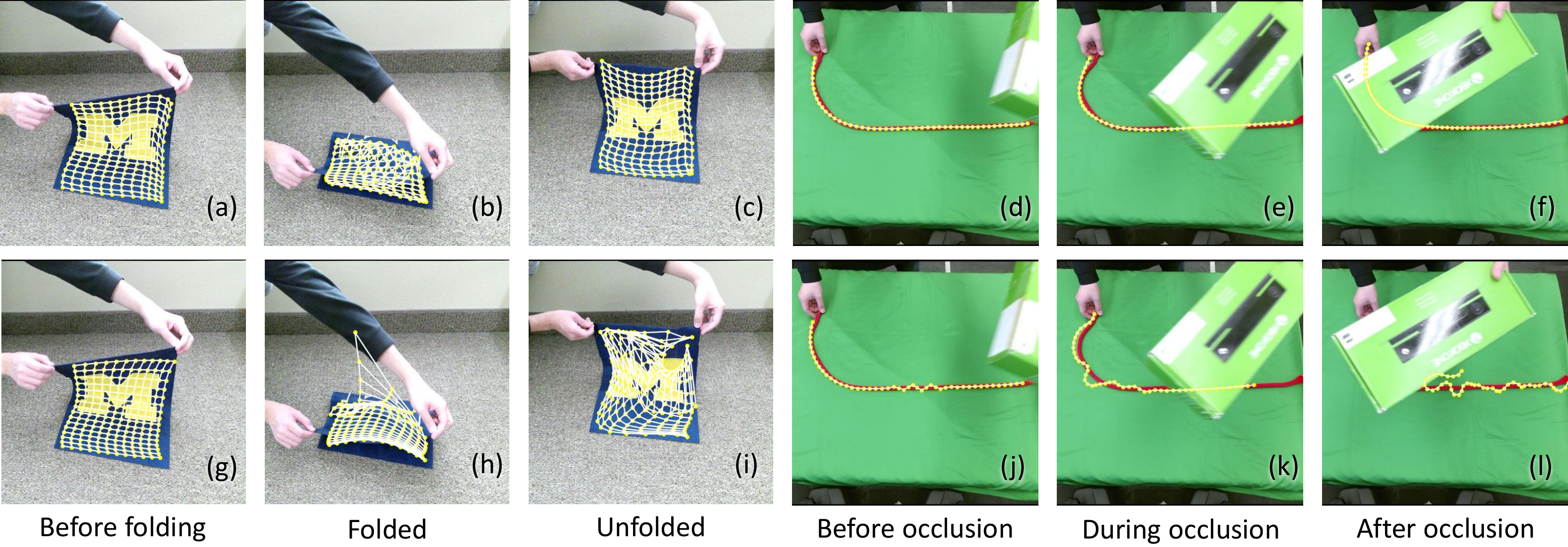}
    \vspace{-0.2in}
    \caption{Results for folding cloth and occluding rope. For cloth, tracking failure recovery engaged for our algorithm after folding. For rope, our algorithm (top row) is not sensitive to the occlusion while CPD+Physics (bottom row) is.}
    \label{fig:cloth}
    \vspace{-.15in}
\end{figure*}

We conducted several experiments tracking rope and cloth to test the performance of our algorithm both quantitatively and qualitatively. These experiments, using both simulation and real-world data, focused on demonstrating the improved robustness against occlusion, as compared to CPD+Physics \cite{robot:TangState}, and the original CPD algorithm \cite{ml:CPD}.

Across all data sets and all three algorithms, we used $\lambda=3.0$, $\beta=1.0$, $\gamma=1.0$, $\tau=0.7$, $k_{vis}=10.0$, $k_{free}=100.0$, and $\epsilon=0.0001$. Point clouds of all rope data sets were down-sampled to $300$ points, and cloth data set were down-sampled to $600$ points. All masks were generated by color filtering. Our algorithm\footnote{An implementation of our algorithm is available on GitHub https://github.com/UM-ARM-Lab/cdcpd/tree/v0.1.0} and original CPD were implemented in python, while CPD+Physics, which is originally implemented in C++, was bridged to python code using the ctypes package. All algorithms were tested on an Intel i7-6700 @ 3.4GHz and 16~GB of RAM.


\begin{figure}[t]
    \centering
    \includegraphics[width=.47\linewidth]{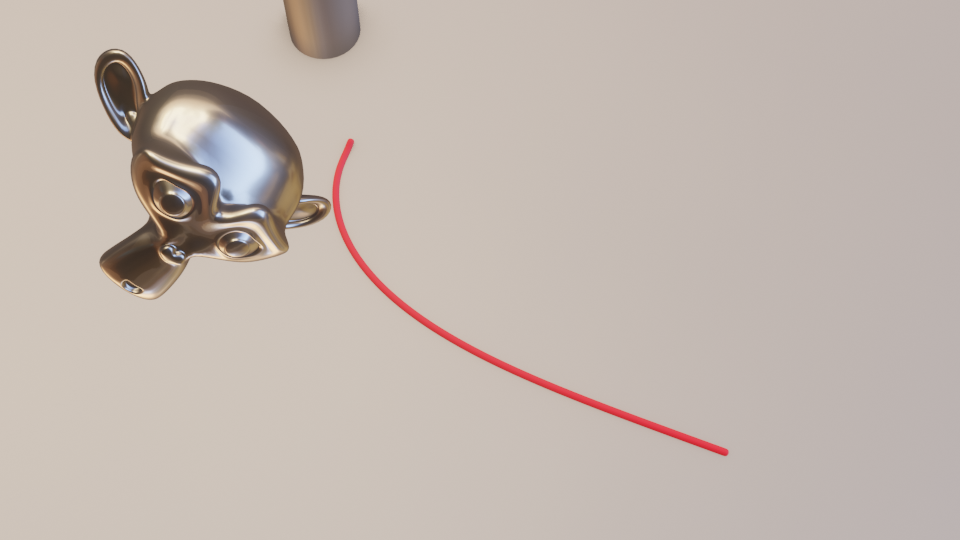}
    \includegraphics[width=.47\linewidth]{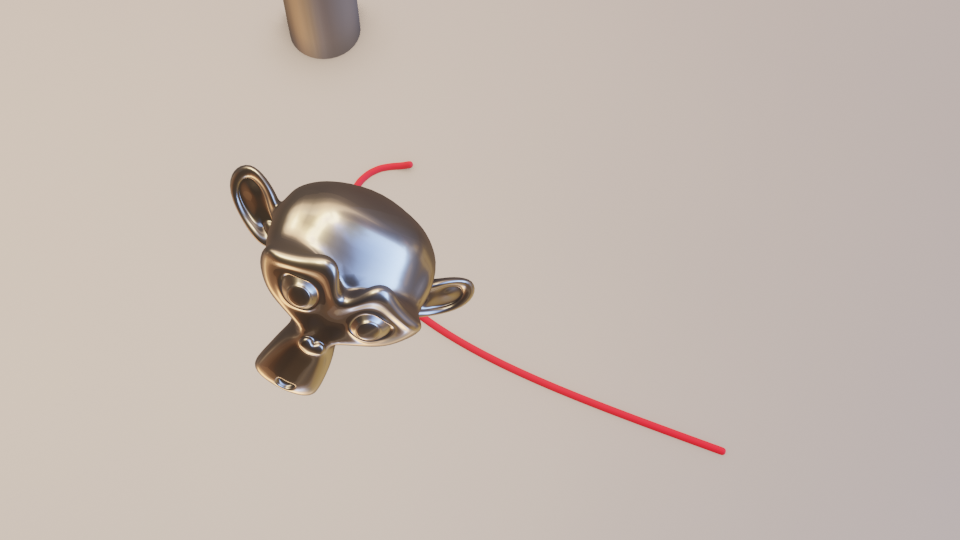}\\
    \vspace{0.05in}
    \includegraphics[width=.47\linewidth]{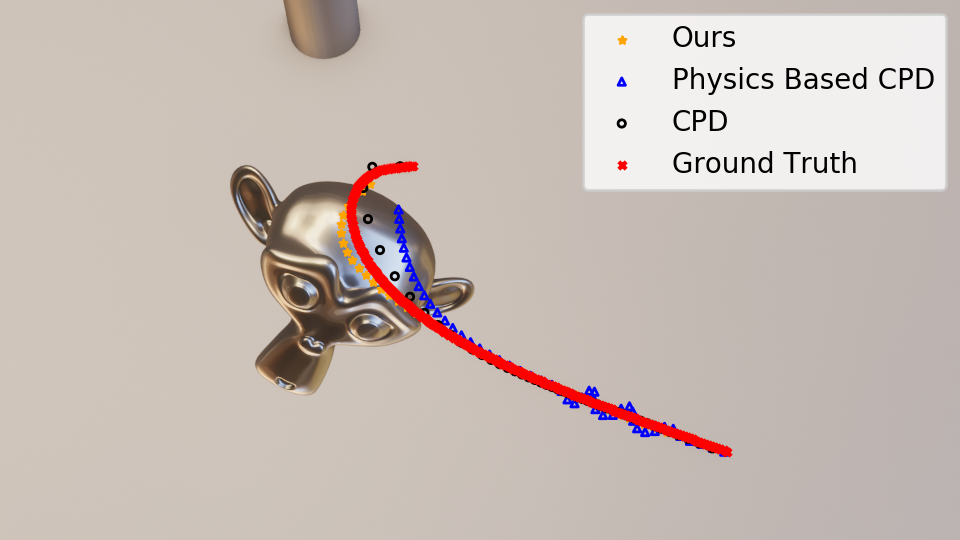}
    \includegraphics[width=.47\linewidth]{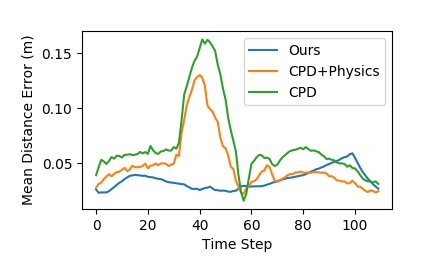} \\
    \caption{Simulated rope experiment. Note that the rope continues moving during occlusion. (TL),(TR) Before and during occlusion. (BL) Tracking output of our method, CPD+Physics, and CPD. (BR) Plot of mean distance error of all three methods.}
    \label{fig:syn_rope}
\end{figure}

\vspace{-.05in}
\subsection{Experiments with Simulated Data}
\vspace{-.02in}
Since the ground truth state of deformable objects is difficult to obtain, we decided to conduct quantitative experiments with a synthetic dataset generated by simulation. We created a virtual table-top with a red rope in 3D modeling software Blender. To compare with CPD+Physics, we tested using a 1-meter-long rope, allowing us to use the same parameters used in \cite{robot:TangState}. In the Bullet simulator used by CPD+Physics the virtual rope was modeled with 50 linked capsules with density $1.5 \rev{\si{\gram\per\centi\meter\cubed}}$, joint stiffness $0.5  \rev{\si{\newton\meter\per\radian}}$, stiffness gain $K_p=10 \rev{\si{\newton\per\meter}}$, and damping gain $K_D=0.5 \rev{\si{\newton\second\per\meter}}$. In the CPD+Physics method, a CPD registration was executed after a each RGBD image arrived, yet the physics simulation steps were executed continuously until the next RGBD image arrived. We executed the physics simulation with a constant 100 steps per frame to fit with our testing framework. 

In the test the rope is dragged by one end with a virtual manipulator (Fig. \ref{fig:syn_rope}(TL)). The rope is approximated with a 48 segment NURBS path and simulated using Blender's internal physics engine. A floating metallic object moves along a predefined path, acting as an occlusion (Fig.~\ref{fig:syn_rope}(TR)). The scene is rendered with EEVEE render engine as 960 $\times$ 540 RGB images, with additional z-buffer output acting as a virtual depth image. A point cloud is then reconstructed using the virtual camera parameters. A Gaussian noise with a standard deviation of $0.002\rev{\si{\meter}}$ was added to the virtual depth image to better approximate real-world sensors. The error plot (Fig. \ref{fig:syn_rope}(BR)) is generated by executing each algorithm 10 times on the data set, and recording the mean of error for each time step across each run. Error is measured by the mean distance of a vertex and corresponding ground truth position.

As seen in Fig. \ref{fig:syn_rope}(BR), our algorithm is less affected by the introduction of occlusion, as compared to the other two algorithms. Between time steps $30$ and $60$, the time span of occlusion, a clear spike in error is seen for the two comparison algorithms, yet our algorithm is less affected. In Fig. \ref{fig:syn_rope}(BL), we can see that the tracking result for CPD around the occluded region is very sparse, and bunched points are visible in the lower right corner for CPD+Physics, indicating that, using these two algorithms,points are "repelled" by the occluded object toward the visible region. The improved performance of our algorithm comes from our visibility-informed membership prior $p_{vis}(m)$. \rev{At around the 55th time step, CPD's tracking result on the previously-occluded part of the rope is over-stretched. The distance constraint in our algorithm pulls vertices closer together, yielding better consistency but also slightly-higher mean squared error.}

\vspace{-.05in}
\subsection{Experiments with Real Data}
\vspace{-.02in}
An experiment of a human manipulating a rope was conducted to demonstrate our algorithm's robustness to occlusion. While continuously moving the rope, the human also deliberately waved a green box in front of the rope, creating a changing occlusion. Fig. \ref{fig:cloth}(d-f) shows three frames of our tracking results while the box moved from right to left. The tracking results in Fig. \ref{fig:cloth}(e-f) maintained the shape of the rope despite occlusion. However, Fig. \ref{fig:cloth}(k-l) shows that the tracking results of CPD+Physics is heavily significantly disrupted by the occlusion.

\begin{figure}[t]
    \centering
        \includegraphics[width=.45\linewidth]{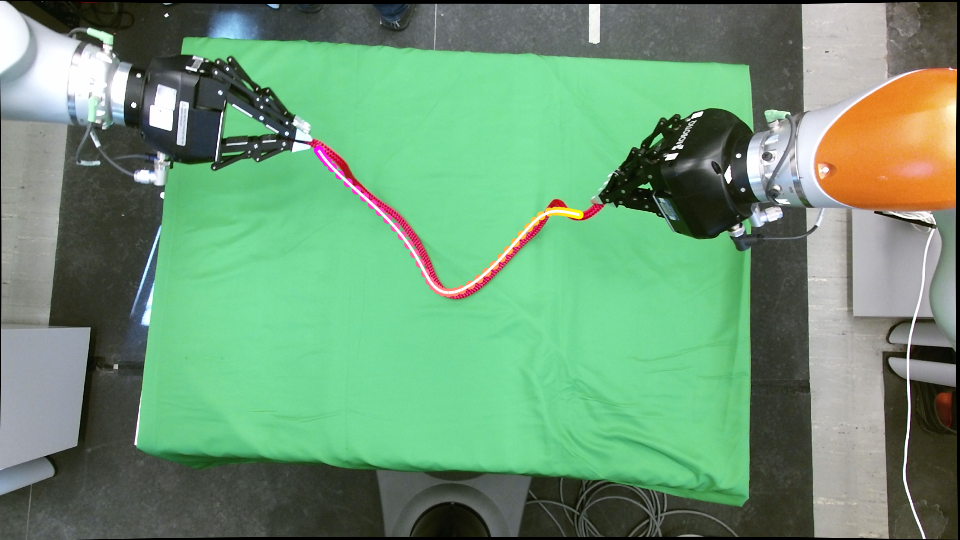} 
        \includegraphics[width=.45\linewidth]{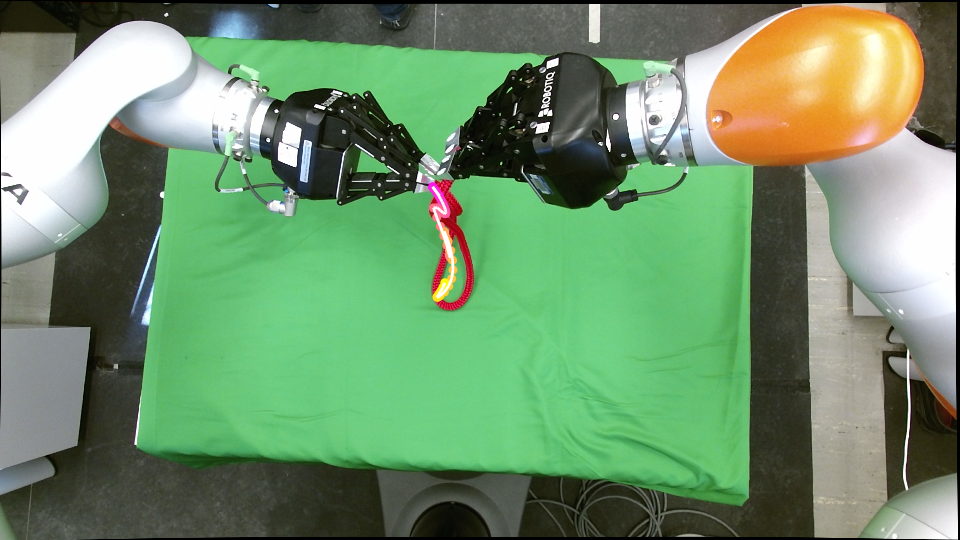}\\
        \vspace{0.05in}
        \includegraphics[width=.45\linewidth]{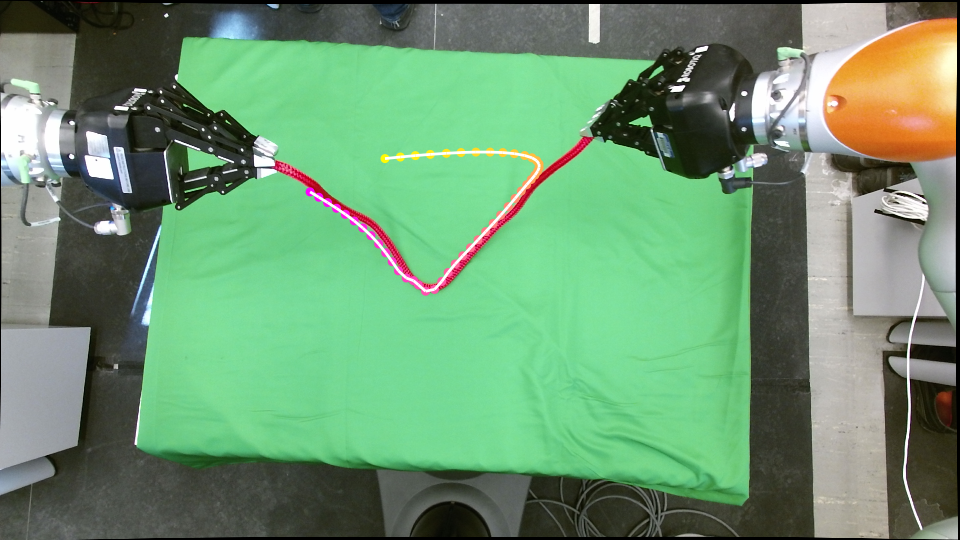} 
        \includegraphics[width=.45\linewidth]{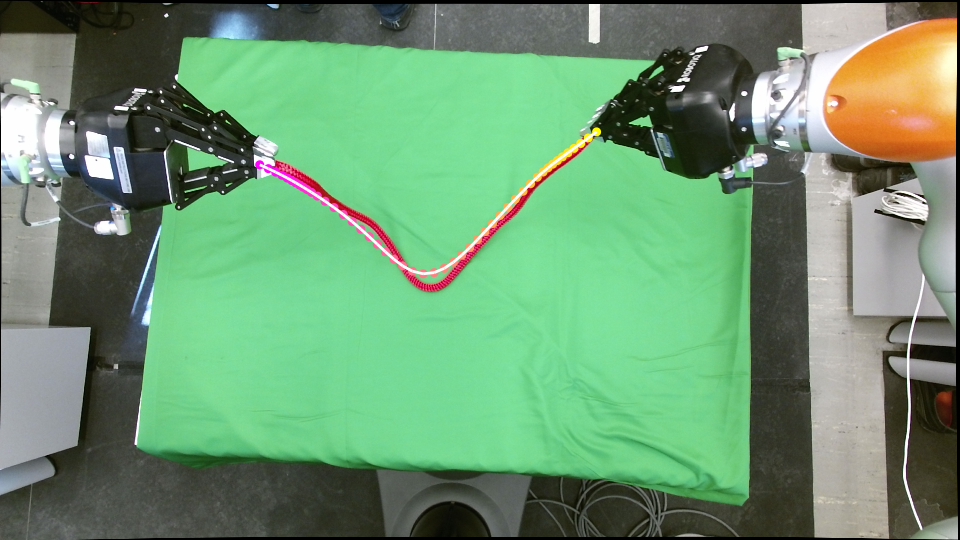} 
    \caption{Results for the overlapping rope test. Vertices have been color-coded to show correspondences. Red and yellow vertices correspond to the left and right half of the rope, respectively, in the initial state (TL). (TR) Our tracking result during overlap. (BL) Output of CPD+Physics after overlap. (BR) Output of our method using gripper correspondences after overlap.}
    \label{fig:gripper}
    \vspace{0.01in}
\end{figure}\hspace{0.05in}
We also tested tracking rope as manipulation was performed by our dual-arm robot to gauge the effect of gripper constraints. The robot folded the rope such that the left and right half overlap with each other. In this case, differentiating the left and right half of the rope is difficult (Fig. \ref{fig:gripper}(TR)). The location of gripped points was calculated using forward kinematics and used in the constrained optimization. As shown in Fig. \ref{fig:gripper}(BL), CPD+Physics failed to track the rope because it chose incorrect correspondences for the endpoints when the rope was overlapping. Our algorithm correctly tracked the rope  (Fig. \ref{fig:gripper}(BR)) due to the inclusion of known correspondences for gripper points.

Our algorithm also generalizes to cloth. Here we compare our method only to the original CPD algorithm because Physics+CPD is limited to only use rope. We conducted an experiment with a human manipulator folding and unfolding a piece of cloth (Fig. \ref{fig:cloth}). Cloth is inherently harder to track, due to more significant self-occlusion. From Fig. \ref{fig:cloth}(i), it is clear that the original CPD algorithm failed to recover tracking after folding and unfolding. Our algorithm exhibits a similar defect as the cloth starts to unfold. However, the tracking failure in our algorithm was successfully detected and tracking was recovered using the tracking result of a previous frame (Fig .\ref{fig:cloth}(c)).

\vspace{-.05in}
\subsection{Computation time}
\vspace{-.02in}
We evaluated the speed of our algorithm on both rope and cloth data sets. We tracked each data set with three tracking failure thresholds $\tau=[0.0, 0.7, 1.0]$, where $\tau=0.7$ represent a typical threshold of tracking failure. When $\tau=1.0$, our algorithm will treat every frame as successful tracking (best case), when $\tau=0.0$ every frame is considered a tracking failure (worst case). We used $k=12$ for the kNN query. Execution time for all major components of our algorithm has been shown in \autoref{table:OursDetailed}. Our algorithm has shown real-time performance in the rope data set when tracking failure occurs infrequently. We also achieved adequate speed for robotic manipulation in the cloth data set.

We also compared the speed of our algorithm with $\tau =0.7$ vs. CPD+Physics and CPD on the human-manipulated rope data. Our method achieves a $43$ms runtime on average, while CPD takes $20$ms and our implementation of CPD+Physics takes $129$ms. Our implementation of CPD+Physics is slower than the reported speed in \cite{robot:TangState} of $20$ms, which is due to the overhead of the ctypes library and frequent memory copy between python and C++ data structures. An implementation of our method in C++ would likely achieve faster runtimes.


\vspace{-.05in}
\subsection{Delicate Motion}
\vspace{-.02in}

While our algorithm demonstrated improved tracking with respect to occlusion, we found that it did not perform well when tracking delicate motions such as tying a knot. In contrast, CPD+Physics tends to perform better for these kinds of motions because a physics simulation allows self-collision checking to regularize the tracking result. In a experiment with data provided by Tang et al., our algorithm was only able to track up until frame 176, while CPD+Physic was able to track all 362 frames. With our method, when a knot is being tied, the rope could interpenetrate, allowing the knot to deteriorate into a straight line. We are currently considering ways to add self-collision information to our method.


\balance

\vspace{-.11in}
\section{Conclusion}
\vspace{-.02in}
We proposed an algorithm that tracks a deformable object without physics simulation. We addressed the occlusion problem by adding a visibility-informed membership prior to our GMM, using constrained optimization for post-processing, and recovering from failed tracking by leveraging free-space information. Our experiments suggest that we improved robustness to occlusion as compared to CPD+Physics and the original CPD algorithm on both simulated and real-world data. In future work, we seek to integrate 3D reconstruction methods to generate a model of the deformable object online and to generalize to cases where the topology of the deformable object changes (e.g. cutting).

\begin{table}[t]
\vspace{5pt}
\caption{Execution Time for Each Components in Our Method}
\label{table:OursDetailed}
\begin{tabularx}{\linewidth}{X|c|X|X|X|X|c}
\hline
                       & $\tau$ & Pre-Proc. & CPD  & Gurobi & Recovery & FPS \\ \hline
\multirow{3}{*}{Rope}  & 1.0    & 10ms     & 11ms & 16ms   & 4ms      & 24  \\ \cline{2-7} 
                       & 0.7    & 10ms     & 11ms & 16ms   & 9ms      & 21  \\ \cline{2-7} 
                       & 0.0    & 9ms      & 14ms & 12ms   & 263ms    & 3   \\ \hline
\multirow{3}{*}{Cloth} & 1.0    & 14ms     & 27ms & 39ms   & 12ms     & 12  \\ \cline{2-7} 
                       & 0.7    & 14ms     & 27ms & 40ms   & 19ms     & 10  \\ \cline{2-7} 
                       & 0.0    & 14ms     & 34ms & 41ms   & 465ms    & 2   \\ \hline
\end{tabularx}
\end{table}









\bibliographystyle{IEEEtran}
\bibliography{references}

\end{document}